# Leveraging LLMs for Predicting Unknown Diagnoses from Clinical Notes

**Dina Albassam[1], Adam Cross[2] and Chengxiang Zhai [1]**
[1]University of Illinois at Urbana-Champaign
[2]University of Illinois, College of Medicine Peoria, Department of Research Services, Peoria, IL, USA.

Correspondence to Adam Cross, PhD, 530 NE Glen Oak Ave Peoria IL 61637; Email: arcross@uic.edu

## Abstract

**Background:** Electronic Health Records (EHRs), including datasets like MIMIC-IV, often lack explicit links between medications and diagnoses, complicating clinical decision-making and research efforts. Even when such links are present, diagnosis lists can be incomplete or inaccurate, particularly during early patient visits when diagnostic uncertainty is high. Discharged summaries, documented at the end of patient care, may offer more detailed explanations of patient visits, potentially aiding in inferring the most likely accurate diagnoses for prescribed medications, especially if we can exploit Large Language Models (LLMs). LLMs have shown promise in processing unstructured medical text, but systematic evaluations are necessary to determine their effectiveness in extracting meaningful medication-diagnosis relationships.

**Objective:** This study explores the use of LLMs to predict implicitly mentioned diagnoses from clinical notes and link them to corresponding medications. We evaluate their effectiveness and investigate strategies to improve prediction performance. Specifically, we examine two research questions: (1) Does majority voting across diverse LLM configurations enhance diagnostic prediction accuracy compared to the best single-model configuration? (2) How sensitive is the diagnostic prediction accuracy of majority voting to the LLM's hyperparameters, including temperature, top-p, and clinical note summary length?

**Methods:** A new dataset of 240 expert-annotated medication-diagnosis pairs from 20 MIMIC-IV clinical notes was created to evaluate predictive accuracy, as no such dataset previously existed. We hypothesized that combining deterministic, balanced, and exploratory configurations could enhance prediction performance. Key hyperparameters—temperature, top-p, and summary length—were systematically varied. Two levels of summarization, short and long, were tested to assess context length impact. Using GPT-3.5 Turbo, 18 configurations were generated, and random subsets of five were selected, resulting in 8,568 test cases. Majority voting was applied to select the most frequent diagnosis. Performance was evaluated using accuracy scores, comparing majority voting with the best single-model configuration and analyzing hyperparameters contributing to the highest accuracy.

**Results:** The majority voting achieved 75% accuracy, outperforming the best single configuration (66%). No single parameter setting consistently excelled; instead, combining diverse configurations aligned with deterministic, balanced, and exploratory strategies yielded better performance. Shorter summaries (2000 tokens) generally improved accuracy. Longer summaries (4000 tokens) were effective only with deterministic settings.

**Conclusions:** Majority voting across LLM configurations enhances diagnostic prediction accuracy in EHRs, demonstrating the potential of ensemble methods for improving medication-diagnosis associations. By leveraging diverse configurations, this approach mitigates model biases and improves robustness in predictive analytics. Future work should explore scalability with larger datasets, additional LLM architectures, and broader clinical applications to refine its effectiveness in real-world settings.

## Introduction

### Background

Accurately linking medications to diagnoses within Electronic Health Records (EHRs) is crucial for improving patient care, supporting clinical decision-making, and advancing predictive analytics. However, this task is challenging due to the lack of explicit connections between medications and diagnoses in EHRs. Even when links exist, diagnosis lists are often incomplete or inaccurate, primarily due to diagnostic uncertainty in the early stages of patient visits [1]. For example, a study found that only 78% of coded diagnoses in EHRs matched the clinical notes, with significant variation based on the documentation workflow. When physicians combined diagnosis coding with note composition, the agreement rate was higher (87.9%) compared to when these tasks were performed separately (44.4%) [2]. Discharged summaries, documented at the end of patient care, may offer more detailed explanations of patient visits, potentially aiding in inferring the most likely diagnoses for prescribed medications, although their completeness can vary. Addressing these challenges requires innovative computational methods to reliably and efficiently establish medication-diagnosis associations.

Recent advancements in Large Language Models (LLMs) have demonstrated their potential in extracting and synthesizing information from unstructured text, such as clinical notes [3]. However, it remains unclear how effectively LLMs can predict diagnoses from clinical notes and associate them with a patient's prescribed medications. In this study, we explore the use of LLMs for predicting unknown diagnoses from clinical notes and linking them to corresponding medications. We study two key research questions: (1) Does majority voting across diverse LLM configurations improve diagnostic prediction accuracy compared to the best single-model configuration? (2) How sensitive is the diagnostic prediction accuracy to the LLM's hyperparameters, including temperature, top-p, and summary length?

As there does not exist any dataset for studying these questions, we leveraged the MIMIC-IV (Medical Information Mart for Intensive Care) dataset [4] to create a new expert-annotated dataset of 240 medication-diagnosis pairs from 20 clinical notes. Using this new dataset, we evaluate the impact of hyperparameter variations and majority voting across multiple LLM configurations on diagnostic prediction accuracy. We first investigate how different configurations of GPT-3.5 Turbo—varying parameters such as temperature, top-p, and summary length—affect diagnostic prediction accuracy. Then, we propose an ensemble approach using majority voting, hypothesizing that pooling diverse LLM configurations from deterministic, balanced, and exploratory groups improves predictive performance.

Our results show that the ensemble approach significantly improves diagnostic prediction accuracy, achieving 75% compared to 66% for the best single-model configuration. This performance boost was achieved by combining diverse configurations from deterministic, balanced, and exploratory strategies, rather than relying on any single optimal parameter setting. Shorter summaries (2000 tokens) consistently enhanced accuracy, while longer summaries (4000 tokens) were effective only when paired with deterministic settings. Additionally, our systematic analysis of hyperparameter sensitivity revealed key interactions that impact prediction accuracy. By leveraging hyperparameter tuning and majority voting, this study enhances LLM-based predictive models and introduces a new benchmark dataset for evaluating medication-diagnosis associations in EHRs. These findings provide practical insights to improve clinical decision-making and support better patient care through more accurate and reliable diagnostic predictions.

## Related Work

### Linking Medications to Diagnoses in EHRs

Several studies have addressed the challenge of linking medications to diagnoses within EHRs. Mullenbach et al. developed a machine learning approach to construct problem-oriented medical records by grouping related medications, procedures, and tests [5]. Another study identified 'medication anomalies' by detecting semantic inconsistencies between prescriptions and diagnoses [6]. Gunasekar et al. utilized Collective Matrix Factorization to integrate diverse EHR sources, uncovering clinically relevant phenotypes [7]. Cole et al. leveraged primary care EHR data to identify patients on specific medications and detect adverse drug events, highlighting EHRs' role in medication safety monitoring [8].

Recent research explores unstructured clinical notes for predicting medications and diagnoses. Yang et al. used a convolutional neural network to analyze admission notes and predict discharge medications, capturing semantic patterns from noisy text [9]. Li et al. applied a similar approach to admission notes for predicting discharge diagnoses, demonstrating the utility of unstructured text in diagnostic decision-making [10]. These studies highlight the value of unstructured clinical notes in enhancing medication-diagnosis linkages.

While these studies provide valuable insights, they either rely on structured EHR data or apply deep learning to specific use cases without explicitly linking medications to diagnoses. Furthermore, no study has leveraged Large Language Models (LLMs) to establish comprehensive medication-diagnosis associations from unstructured clinical notes. Our study addresses these gaps by using LLMs to extract medication-diagnosis relationships from unstructured notes.

*Leveraging Large Language Models (LLMs) for Medical Knowledge Extraction and Retrieval*

Advances in LLMs have significantly influenced the medical field, particularly in knowledge extraction and retrieval from unstructured clinical text. For instance, Gu et al. explored task-specific distillation of LLMs for biomedical applications, achieving substantial improvements in structuring text for adverse drug event extraction [11]. Similarly, Wang et al. demonstrated the potential of fine-tuned LLaMA models for clinical note analysis, leading to improved Diagnosis-Related Groups (DRG) prediction performance on the MIMIC-IV dataset [12]. Modular approaches, such as the integration of search engine modules, have also enhanced the factual accuracy of medical information retrieval systems [13].

Moreover, frameworks evaluating LLM performance against medical benchmarks have emerged to assess clinical knowledge in models like ChatGPT [3]. These frameworks underline the need for task-specific fine-tuning and human evaluations to ensure clinical accuracy. Innovations in zero-shot natural language generation highlight the potential for automating tasks like clinical documentation [14], while educational applications of LLMs have revealed their value in healthcare staff training [15]. Recently, domain-specific models such as Med-PaLM 2 have set benchmarks for expert-level question answering, underscoring the importance of optimizing LLMs for healthcare [16].

Given the significant potential of LLMs for medical knowledge extraction, our study utilizes GPT-3.5 Turbo with zero-shot prompting to enhance medication-diagnosis linkage from unstructured clinical notes.

*Ensemble Methods and Diversity in LLM Configurations*

Ensemble techniques like majority voting have emerged as powerful tools for improving LLM performance, especially in domains requiring high accuracy and robustness. Techniques such as EnsemW2S leverage collaborative weak learners to enhance stronger models, achieving notable accuracy improvements in sparse data scenarios [17]. LLM-TOPLA further optimizes ensembles by introducing diversity metrics and pruning algorithms, achieving state-of-the-art performance on tasks like MMLU and SearchQA [18].

In healthcare, ensembles such as LLM-Synergy have shown effectiveness in medical question-answering, using weighted majority voting and dynamic model selection to surpass individual model accuracy on datasets like MedQA-USMLE and PubMedQA [19]. Additionally, reasoning with LLMs for medical question answering has been enhanced through ensemble reasoning techniques, improving consistency and accuracy by refining reasoning processes [20]. Studies also highlight the importance of diversity in ensemble configurations, varying parameters such as temperature, top-p, and model architecture to mitigate biases and improve generalization [17,18]. These methods address the inherent variability in LLM outputs, showcasing the value of ensemble approaches in producing reliable predictions.

Although ensemble methods have been effectively applied in healthcare for question answering and knowledge extraction, no study has explored their potential in linking medications to diagnoses from unstructured clinical notes. Furthermore, existing studies do not investigate hyperparameter sensitivity or leverage a systematic approach to varying configurations across deterministic, balanced, and exploratory strategies. Our study introduces an innovative ensemble approach using majority voting across diverse LLM configurations, systematically analyzing hyperparameter sensitivity to optimize model performance.

## Methods

We address the challenge of predicting the most likely diagnosis for each medication in clinical records, a problem complicated by the lack of annotated datasets with ground truth. To tackle this, we leverage large language models (LLMs) and hypothesize that combining predictions from multiple diverse configurations using majority voting can enhance diagnostic prediction accuracy. By systematically varying parameters such as temperature, top_p, and clinical note summary length, we create a diverse set of configurations. We then evaluate the best-performing subsets of five configurations to analyze their contributions to prediction performance and gain insights into parameter interactions that drive ensemble success. (Figure 1) illustrates this workflow, detailing the dataset preparation, configuration selection, diagnostic prediction, ensemble testing, and performance evaluation process.

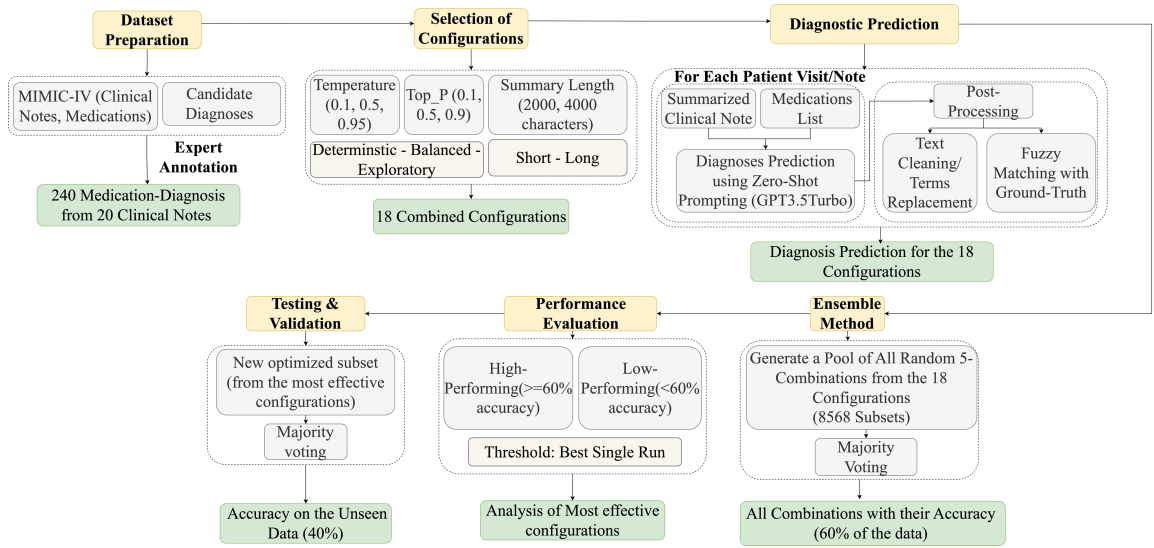

**Figure 1:** *Workflow for Diagnostic Prediction and Ensemble Method*

## Dataset Description

We utilized the Medical Information Mart for Intensive Care (MIMIC-IV) dataset [4], a large-scale, de-identified dataset comprising health-related data from over forty thousand patients who were admitted to the critical care units of the Beth Israel Deaconess Medical Center between 2008 and 2019. MIMIC-IV includes diverse data types, including vital signs, medications, lab measurements, and clinical notes.

For this study, due to the absence of ground truths for direct medication-diagnosis associations, we generated candidate diagnoses using GPT-3.5Turbo from a subset of clinical notes paired with 350 medications. These candidates were then reviewed and annotated by a clinical expert, resulting in a curated list of 240 confirmed medication-diagnosis pairs derived from 20 clinical notes. The process of annotating and reviewing was challenging, as each medication-diagnosis pair was unique to its corresponding clinical note, requiring a thorough review of the note to ensure the accuracy of the annotated data.

## Selection of Configurations to be Tested

The configurations tested in our approach are systematically structured to maximize diversity and capture complementary behaviors in model predictions. As shown in (Figure 1), the configurations are categorized based on three primary parameters:

The temperature parameter controls the randomness of predictions, influencing how deterministic or exploratory the model's responses are [21]. A deterministic setting (0.1) produces highly consistent and predictable outputs with minimal randomness, ensuring stability in predictions. A balanced setting (0.5) maintains a middle ground, allowing some degree of exploration while retaining control over the output. On the other hand, an exploratory setting (0.95) generates highly diverse and random predictions, increasing variability and favoring the exploration of less likely options.

The Top_p parameter regulates the flexibility in predictions by controlling the probability threshold for token selection [22]. A low setting (0.1) restricts the prediction space to the highest-probability tokens, ensuring precision and a narrow focus. A medium setting (0.5) balances flexibility by incorporating a wider range of probable tokens, leading to more nuanced predictions. Meanwhile, a high setting (0.9) enhances prediction diversity by sampling from a larger pool of tokens, allowing for more varied and potentially novel outputs.

The length of a clinical note summary determines the level of contextual information available to the model. Short summaries (~2000 characters) provide a condensed version, highlighting only the most critical details. In contrast, long summaries (~4000 characters) offer extended context, incorporating a broader range of information that may enhance predictive accuracy. Given that the average clinical note is approximately 6400 characters, these summaries represent varying degrees of detail and completeness.

By combining these parameters, we create a comprehensive set of 18 configurations. Each configuration reflects a unique trade-off between randomness, flexibility, and context, ensuring the ensemble captures diverse perspectives. This structured variation allows us to systematically evaluate their individual and combined contributions to diagnostic prediction accuracy. The analysis of high- and low-performing configurations within this framework enables insights into optimizing ensemble methods for this task.

## Diagnostic Prediction

### Summarization for Context Variations

To examine the influence of context length on prediction accuracy, we summarize clinical notes into shorter (~2000 characters) and longer (~4000 characters) versions. This variation allows the model to evaluate how different levels of detail affect diagnostic performance. GPT-3.5Turbo is utilized for this task with the following prompt:

```
def simplify_medical_note(note):
    gpt_response = process_prompt("Summarize the clinical note,
```

```
     and make its length < summaryLength" + note)
```

By varying the summary length across configurations, we systematically investigate the effect of contextual detail on diagnostic prediction.

*Zero-Shot Prompting*

To predict the most likely diagnosis for each medication within a clinical context, we use a zero-shot prompting approach with GPT3.5Turbo. The model is guided by the following prompt

```
"Given a patient's clinical note: '{clinical_note}', and the
medication: {medication}, what diagnosis is the most likely indication
for this medication in this specific patient? In other words, what
diagnosis is the medication treating in this context? Return the name
of the diagnosis only."
```

This targeted prompt ensures the model focuses on the clinical context, generating concise, relevant, and context-aware predictions. The zero-shot approach allows for flexible and scalable diagnostic prediction without requiring extensive task-specific fine-tuning.

*Data post-processing*

Predicted diagnoses were cleaned and standardized to ensure consistency in text representation. Common shorthand terms and variations of medical diagnoses were replaced with standardized terminology to reduce ambiguity and improve alignment with ground-truth labels. For example, "chf" was expanded to "congestive heart failure," and "gerd" to "gastroesophageal reflux disease." Additionally, non-alphanumeric characters such as slashes ("/") and hyphens ("-") were removed to simplify text parsing. After post-processing, fuzzy matching was applied using a 60% similarity threshold to compare model predictions against annotated ground-truth diagnoses. This step served as a critical measure for evaluating prediction accuracy and establishing an initial performance benchmark.

*Generating Combinations and Applying Majority Voting*

From the 18 predefined configurations of temperature, top_p, and clinical note summary length, random subsets of five combinations were generated to evaluate the ensemble's performance, resulting in 8,568 test cases. These combinations represent diverse perspectives by incorporating different levels of randomness, flexibility, and context. For each combination, predictions were aggregated using a majority voting mechanism, which determines the final diagnostic output based on the most frequently predicted diagnosis across configurations in the subset. The accuracy of majority voting was calculated on the training data (60% of the dataset) by comparing ensemble predictions to the ground truth, identifying which combinations performed best overall.

*Performance Evaluation and Testing*

To evaluate the effectiveness of the ensemble approach, configurations were split based on their performance into best-performing subsets (≥60% accuracy, exceeding the accuracy threshold of 59%, which was the best single-run accuracy) and lower-performing subsets (<60%). A turn-level analysis was then conducted to identify configurations that consistently contributed to the success of high-performing ensembles. This analysis provided insights into the parameter settings that were most effective in diagnostic prediction. Based on the findings, a new combination subset was generated by incorporating frequently occurring high-performing configurations. These optimized combinations were subsequently tested on the unseen test dataset (40% of the data) to validate their performance and assess the generalizability of the approach.

**Accuracy Measurement:** The effectiveness of our approach is primarily evaluated using the accuracy metric. Accuracy is computed by comparing the model's predicted diagnoses against the ground truth diagnoses annotated by clinical experts. Specifically, accuracy is calculated as the proportion of correct predictions out of the total number of predictions made by the model. This metric provides a straightforward measure of how well the LLMs perform in identifying the correct diagnosis from the clinical notes and medications provided.

## Results

In this section, we aim to address the key research questions (RQs) regarding the effectiveness of using Large Language Models (LLMs) and ensemble methods for predicting diagnoses based on clinical notes and medications. The questions are as follows:

(1) Does majority voting across diverse LLM configurations improve diagnostic prediction accuracy compared to the best single-model configuration?

(2) How sensitive is the diagnostic prediction accuracy of majority voting to the LLM's hyperparameters, including temperature, top-p, and clinical note summary length?

We will analyze these questions by examining the performance of different ensemble configurations, exploring their contributions to overall accuracy, and providing insights into the key factors that drive ensemble success. Detailed results, examples, and explanations are presented to highlight the major findings and their implications.

**Performance Analysis**

To address RQ1: Does majority voting across diverse LLM configurations improve the accuracy of diagnostic predictions compared to the best single configuration? We split the data into 60% training and 40% testing. For the training set, we first computed the accuracy of all 18 configurations, identifying the best-performing single run. Next, we generated all possible 5-combinations from these configurations and evaluated their majority voting performance. We then selected the highest-performing combination and compared it to the best single run, as shown in (Table 1) and (Table 2). The results indicate that the best single run achieved 59% accuracy, while majority voting improved accuracy to 75% in the highest-performing combination, with the lowest combination scoring 50%.

Table 1. *Performance of Different Model Configurations for Diagnostic Prediction (Training Data)*

| Configurations | Temperature | Summary Length | Top_P | Performance (%Acc) |
|---|---|---|---|---|
| | | | | |
| Config1 | 0.1 | 2000 | 0.1 | 56% |
| Config2 | 0.1 | 2000 | 0.5 | 55% |
| Config3 | 0.1 | 2000 | 0.9 | 56% |
| Config4 | 0.1 | 4000 | 0.1 | 55% |
| Config5 | 0.1 | 4000 | 0.5 | 54% |
| Config6 | 0.1 | 4000 | 0.9 | 54% |
| Config7 | 0.5 | 2000 | 0.1 | 57% |
| Config8 | 0.5 | 2000 | 0.5 | *59%* |
| Config9 | 0.5 | 2000 | 0.9 | 55% |
| Config10 | 0.5 | 4000 | 0.1 | 53% |
| Config11 | 0.5 | 4000 | 0.5 | 53% |
| Config12 | 0.5 | 4000 | 0.9 | 58% |

| Config13 | 0.95 | 2000 | 0.1 | 58% |
| Config14 | 0.95 | 2000 | 0.5 | *59%* |
| Config15 | 0.95 | 2000 | 0.9 | 57% |
| Config16 | 0.95 | 4000 | 0.1 | 55% |
| Config17 | 0.95 | 4000 | 0.5 | 54% |
| Config18 | 0.95 | 4000 | 0.9 | 52% |

Table 2. Performance Comparison of Best Single Run and Majority Voting (Training Data)

| Method | Performance (%Acc) |
|---|---|
|  |  |
| Best Single Run | 59% |
| Majority Voting (Highest-Accuracy) | 75% |
| Majority Voting (Lowest-Accuracy) | 50% |

## *Turns Leading to High Accuracy*

To address RQ2: How sensitive is the diagnostic prediction accuracy of majority voting to the LLM's hyperparameters, including temperature, top-p, and clinical note summary length? We analyzed the most frequent configurations contributing to high-performing versus low-performing combinations, as illustrated in the bar charts in (Figure 2). Additionally, we examined the agreement between top-performing configurations using a confusion matrix, as shown in (Figure 3).

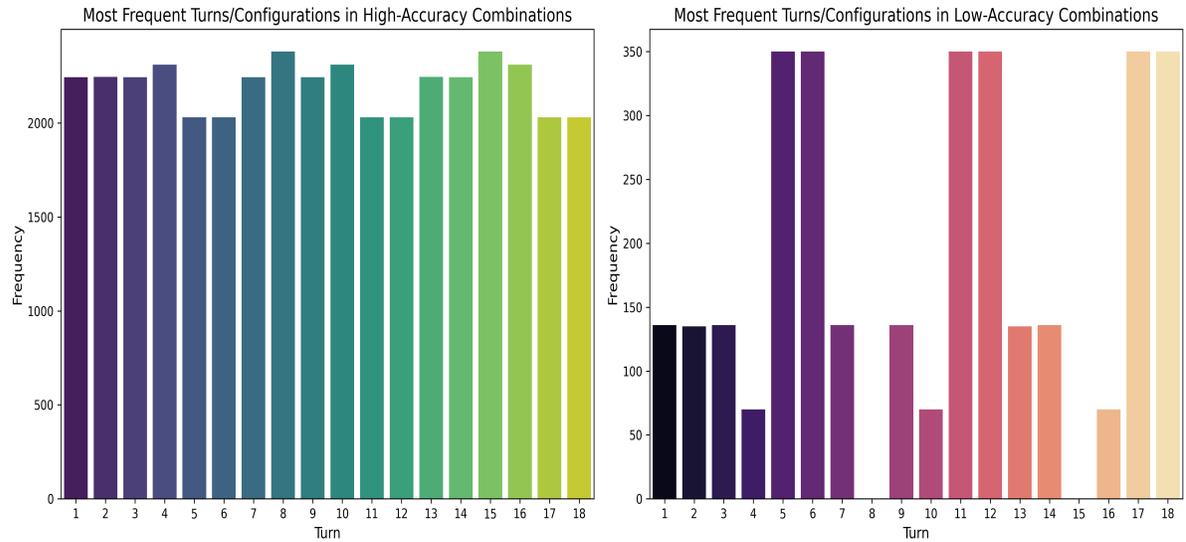

**Figure 2: Most Frequent Turns/Configurations in High- and Low-Accuracy Combinations**
This comparison of bar plots shows the frequency of individual turns across high-accuracy (left) and low-accuracy (right) configurations. In high-accuracy combinations, turns are more evenly distributed, suggesting balanced contributions from multiple configurations. Conversely, low-accuracy combinations exhibit a concentration of specific turns, indicating potential dependencies on less effective configurations. This analysis informs the selection of optimized subsets for majority voting.

As shown in the bar chart, the most frequent turns (configurations) in high-accuracy combinations include 1, 2, 3, 4 (deterministic), 7, 8, 9, 10 (balanced), and 13, 14, 15, 16 (exploratory). These turns span across all three configuration categories for the temperature parameter, highlighting the importance of parameter diversity in ensemble methods.

A key observation is that while shorter summaries (2000 tokens) consistently yield optimal performance across various configurations—whether deterministic, balanced, or exploratory—longer summaries (4000 tokens) perform well only when at least one of the parameters (temperature or top-p) is set to a low (deterministic) value. When both parameters are high (balanced or exploratory), longer summaries tend to result in lower performance, as reflected in the bar chart (Figure 2), which illustrates the combinations associated with low accuracy. This suggests that excessive variability in generation settings may negatively impact majority voting outcomes when longer summaries are used.

### Intersection of Turns in Top Combinations

The analysis of intersections among the top 10-performing combinations (accuracy = 75%) reveals that the most commonly agreed-upon turns are 2, 7, 10, 13, and 14.

These turns highlight the complementary roles played by configurations spanning deterministic, balanced, and exploratory categories for the temperature parameter offering insights into the design of effective ensembles.

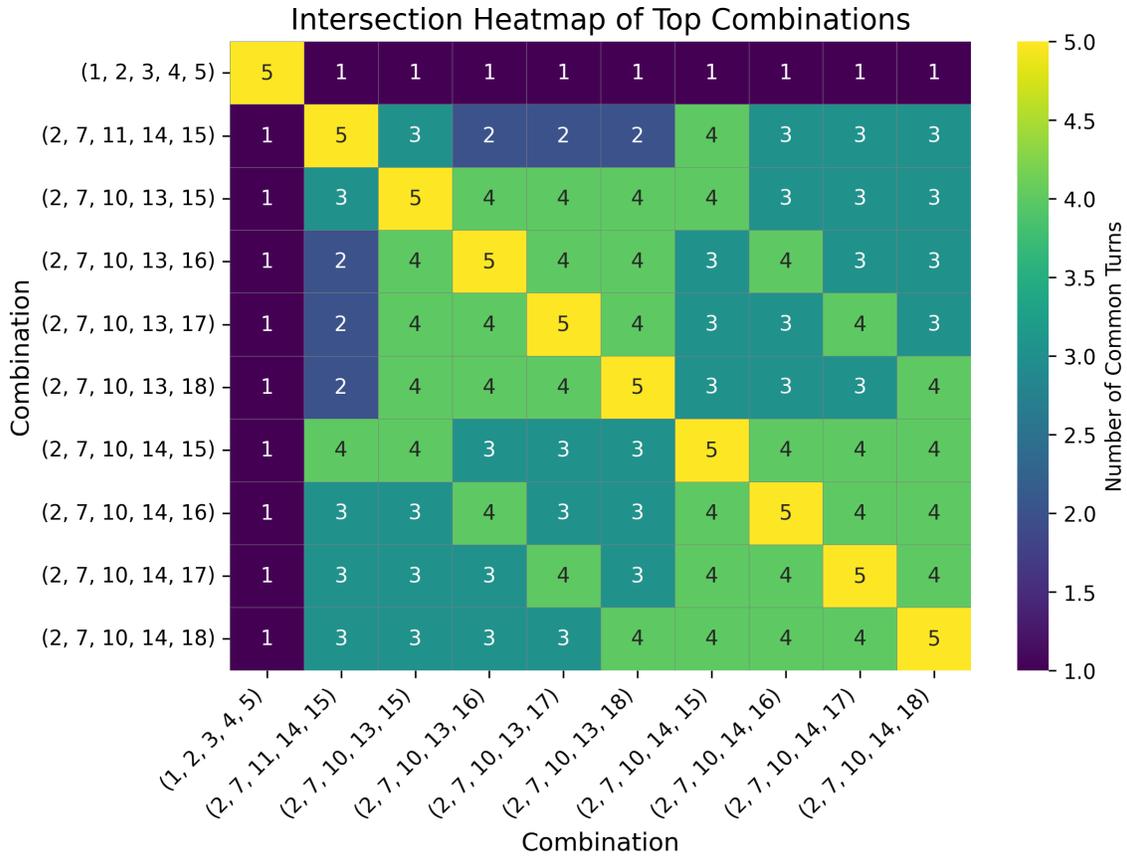

**Figure 3: Intersection Heatmap of Top Combinations**
This heatmap illustrates the intersections among the top 10 configurations with the highest majority voting accuracy. Each cell represents the number of common turns between pairs of combinations. The discrete color scale (1-5) highlights the degree of overlap, revealing patterns of shared configurations that contribute to high predictive accuracy. The diagonal values indicate the total number of turns within each combination.

### Agreed Turns Across Top Combinations

The agreed turns consistently present in the top-performing combinations highlight the complementary roles of deterministic, balanced, and exploratory configurations. Turn 2 (Deterministic: 0.1, 2000, 0.5) provides a stable and consistent foundation for predictions due to its deterministic nature and concise summaries. Turn 7 (Balanced: 0.5, 2000, 0.1) strikes an effective balance between exploration and stability, enabling the ensemble to capture key patterns while maintaining precision. Turn 10 (Balanced: 0.5, 4000, 0.1) expands the context slightly with a longer summary length

(4000 tokens) while preserving stability through balanced temperature settings. Turn 13 (Exploratory: 0.95, 2000, 0.1) introduces diversity by allowing higher randomness in predictions, with the shorter summary length (2000 tokens) helping to focus the randomness on critical details and promoting a broader search space for potential diagnoses. Finally, turn 14 (Exploratory: 0.95, 2000, 0.5) balances exploratory randomness with moderate flexibility (medium top_p), allowing the ensemble to generalize better and uncover patterns that deterministic and balanced configurations might overlook.

To demonstrate how these configurations complement each other, (Table 3) illustrates the effectiveness of majority voting in enhancing diagnostic prediction accuracy compared to the best single configuration. The majority voting approach aggregates predictions from five diverse configurations (Turns 2, 7, 10, 13, and 14), representing deterministic, balanced, and exploratory parameter settings. This diversity allows the ensemble to capture different aspects of the clinical context, mitigating the limitations of individual configurations.

For instance, in the case of Enalapril Maleate prescribed for hypertension, majority voting correctly identified the diagnosis by combining the accurate predictions from Turns 7, 10, and 13, despite Turn 14 failing to make the correct prediction. Similarly, for Ondansetron associated with gastric distension, the majority voting approach successfully identified the diagnosis by leveraging the correct outputs from Turns 2, 10, and 13.

Table 3. Cases Where Majority Voting Outperformed the Best Single Run

| Medication | Diagnosis | Best Run (Turn 14) | Turn 2 | Turn 7 | Turn 10 | Turn 13 | Turn 14 |
|---|---|---|---|---|---|---|---|
| | | | | | | | |
| Enalapril Maleate | ['hypertension'] | 0 | 0 | 1 | 1 | 1 | 0 |
| Ondansetron | ['gastric distension'] | 0 | 1 | 0 | 1 | 1 | 0 |

*Results on Testing Data*

To address RQ1 and demonstrate the advantage of majority voting over single configurations, we evaluated the performance accuracy of all 18 configurations on the testing data, as shown in (Table 4). The best single configuration achieved 66% accuracy. We then compared this to the majority voting ensemble applied to the test data (40% of the dataset), as shown in (Table 4). The majority voting ensemble was formed using the most frequent configurations that contributed to the highest accuracy during training: (15, 8, 10, 16, 2). The results, summarized in (Table 5), show that majority voting outperformed the best single configuration, achieving 75% accuracy compared to 66%.

Table 4. *Performance of Different Model Configurations for Diagnostic Prediction (Testing Data)*

| Configurations | Temperature | Summary Length | Top_P | Performance (%Acc) |
|---|---|---|---|---|
| Config1 | 0.1 | 2000 | 0.1 | 64% |
| Config2 | 0.1 | 2000 | 0.5 | 63% |
| Config3 | 0.1 | 2000 | 0.9 | 63% |
| Config4 | 0.1 | 4000 | 0.1 | 60% |
| Config5 | 0.1 | 4000 | 0.5 | 64% |
| Config6 | 0.1 | 4000 | 0.9 | 63% |
| Config7 | 0.5 | 2000 | 0.1 | 65% |
| Config8 | 0.5 | 2000 | 0.5 | 60% |
| Config9 | 0.5 | 2000 | 0.9 | 65% |
| Config10 | 0.5 | 4000 | 0.1 | 62% |
| Config11 | 0.5 | 4000 | 0.5 | 61% |
| Config12 | 0.5 | 4000 | 0.9 | *66%* |
| Config13 | 0.95 | 2000 | 0.1 | 58% |
| Config14 | 0.95 | 2000 | 0.5 | 64% |
| Config15 | 0.95 | 2000 | 0.9 | 63% |
| Config16 | 0.95 | 4000 | 0.1 | 60% |
| Config17 | 0.95 | 4000 | 0.5 | 63% |

| Config18 | 0.95 | 4000 | 0.9 | 63% |

Table 5. Performance Comparison of Best Single Run and Majority Voting (Testing Data)

| Method | Performance (%Acc) |
|---|---|
|  |  |
| Best Single Run | 66% |
| Majority Voting | 75% |

# Discussion

## Principal Results

This study demonstrates that leveraging Large Language Models (LLMs) for predicting unknown diagnoses from clinical notes and linking them to medications can significantly enhance diagnostic accuracy when using majority voting across diverse configurations. The ensemble approach achieved a 75% accuracy rate, outperforming the best single configuration's accuracy of 66%. The results indicate that no single configuration consistently yielded the highest performance. Instead, combining deterministic, balanced, and exploratory configurations enabled the ensemble to effectively capture diverse perspectives, enhancing robustness and generalization in diagnostic predictions. Furthermore, shorter summaries (~2000 tokens) consistently improved accuracy, while longer summaries (~4000 tokens) were effective only with deterministic settings. These findings suggest that varying contextual detail and prediction randomness are critical in optimizing LLM-based clinical decision support systems.

## Limitations

Despite achieving promising results, this study has several limitations. First, due to the lack of any existing dataset for our study and the limited resources, we were only able to conduct the evaluation on a relatively small dataset of 240 expert-annotated medication-diagnosis pairs derived from 20 clinical notes. While this dataset was meticulously curated, its limited size may restrict the generalizability of the findings. Second, the study utilized a single LLM architecture (GPT-3.5 Turbo) without comparing it to other state-of-the-art models such as GPT-4 or domain-specific models like Med-PaLM 2. Finally, the findings are based on majority voting ensembles, and alternative ensemble methods such as weighted voting or stacking were not explored.

**Implications for Future Research**

This study underscores the potential of majority voting ensembles in enhancing LLM-based diagnostic predictions. Since the diversity-based majority voting strategy is orthogonal to our specific task, the strategy can be potentially used for many other tasks to improve effectiveness of LLMs. It would be interesting to further test such a strategy with other models, different tasks, and larger datasets.  Future research should explore the scalability of this approach using larger and more diverse clinical datasets to validate the generalizability of the findings. Exploring alternative ensemble methods, such as weighted majority voting or stacking, could further optimize prediction accuracy. Comparing different LLM architectures, including domain-specific models like Med-PaLM 2, would also offer insights into model robustness and performance variations. Furthermore, incorporating human-in-the-loop evaluations, particularly from clinical experts, would enhance the clinical validity and usability of the predictions. Finally, investigating the impact of LLM hallucinations and developing mitigation strategies would improve the reliability and safety of LLM-based clinical decision support systems.

**Conclusion**

This study demonstrates the effectiveness of leveraging Large Language Models (LLMs) for predicting unknown diagnoses from clinical notes and linking them to medications. By applying a majority voting approach across diverse LLM configurations, we achieved improved predictive accuracy compared to single-model configurations. This is encouraging because the strategy is general and can be potentially applied on top of any LLM for any prediction task to improve prediction accuracy.  Our findings show that no single parameter configuration consistently yields the best performance; instead, combining deterministic, balanced, and exploratory settings enhances robustness and generalization.

The results highlight that shorter summaries (2000 tokens) are generally more effective across configurations, while longer summaries (4000 tokens) only perform well when paired with a deterministic setting (low temperature or top-p). Excessive randomness in exploratory settings, combined with extended context, tends to reduce predictive accuracy. These insights are crucial for optimizing LLM-based clinical decision support systems, emphasizing the importance of controlled diversity in ensemble models.

Beyond demonstrating the feasibility of LLMs for medication-diagnosis linkage, this study underscores the potential for ensemble methods to mitigate individual model biases and improve reliability in medical AI applications. Our approach lays the foundation for future research on scaling predictive models with larger datasets, integrating additional LLM architectures, and refining ensemble selection strategies.

## Acknowledgements

We would like to thank Professor Jimeng Sun for supporting us with access to OpenAI Microsoft Azure resources.

## Conflicts of Interest

none declared.